\documentclass[letterpaper, 10 pt, conference]{ieeeconf}  %

\IEEEoverridecommandlockouts                              

\usepackage{amssymb}
\usepackage{algorithm}
\usepackage{algpseudocode}
\usepackage{graphicx}
\usepackage[flushleft]{threeparttable}
 \usepackage{amsmath} 
\usepackage{subcaption}
\usepackage{multirow}
\usepackage[noadjust]{cite}

\title{\LARGE \bf
Distracted Robot: How Visual Clutter Undermine Robotic Manipulation
}

\author{
    Amir Rasouli\thanks{Corresponding author\texttt{\{amir.rasouli@huawei.com\}}. Huawei Technologies Canada. $^{\dagger}$Work done while at Huawei Canada. } \quad Montgomery Alban$^{\dagger}$ \quad Sajjad Pakdamansavoji \quad Zhiyuan Li\\ \quad Zhanguang Zhang \quad Aaron Wu \quad Xuan Zhao \\ \\
}

\begin{document}

\maketitle

\thispagestyle{empty}
\pagestyle{empty}


\begin{abstract}
In this work, we propose an evaluation protocol for examining the performance of robotic manipulation policies in cluttered scenes. Contrary to prior works, we approach evaluation from a psychophysical perspective, therefore we use a unified measure of clutter that accounts for environmental factors as well as the distractors quantity, characteristics, and arrangement. Using this measure, we systematically construct evaluation scenarios in both hyper-realistic simulation and real-world and conduct extensive experimentation on manipulation policies, in particular vision-language-action (VLA) models. Our experiments highlight the significant impact of scene clutter, lowering the performance of the policies, by as much as 34\% and show that despite achieving similar average performance across the tasks, different VLA policies have unique vulnerabilities and a relatively low agreement on success scenarios. We further show that our clutter measure is an effective indicator of performance degradation and analyze the impact of distractors in terms of their quantity and occluding influence. At the end, we show that finetuning on enhanced data, although effective, does not equally remedy all negative impacts of clutter on performance.
\end{abstract}

\section{Introduction}
\begin{figure}
    \centering
    \includegraphics[width=0.7\linewidth]{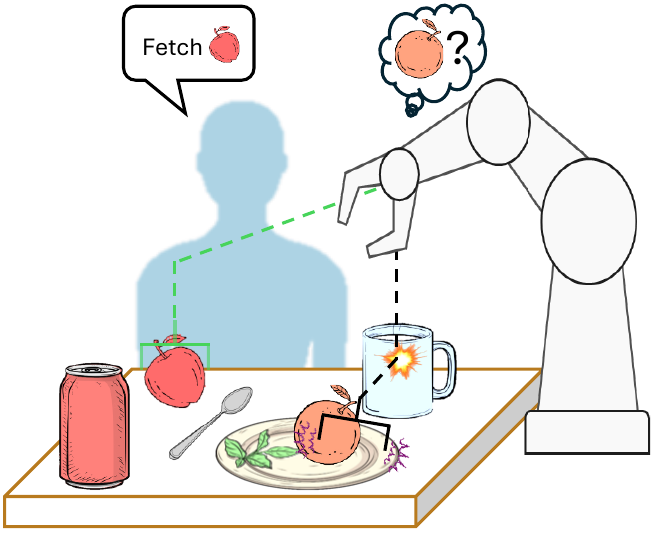}
    \caption{A typical manipulation scenario where the robot is asked to fetch an object (e.g. an apple). The green trajectory shows the expected behavior and black shows the executed one. Distractors in the scene can cause target confusion (e.g. with the orange), as well as collision and grasping failure. }\vspace{-0.4cm}
    \label{fig:first_image}
\end{figure}

One of the key requirements for the deployment of robots in real world is robustness to clutter and variation in environmental characteristics. Past studies show that clutter caused by distractors, i.e. non-targets that do not serve any purpose for accomplishing the tasks, can adversely impact the performance of robotic policies \cite{Pumacay_RSS_24,Yu_RSS_23,mendez2021improving, sun2024causalagents}. Distractors in the scene can potentially lead to misperception due to partial observability resulting from occlusion, target confusion due to semantic or visual similarity, incorrect associations with the task, or cause obstruction (see Figure \ref{fig:first_image}).

Given the importance of context on the success of robotic policies, it is crucial to design effective evaluation protocols to systemically identify the limitations of the policies under different environmental conditions. Most existing evaluation protocols focus on the types of robotic skills, the choice of target objects, and the types of reasoning capabilities given different inputs  \cite{liu2023libero,zhu2020robosuite,mayoral2024robotperf}. These works do not specifically look at the effect of distractors, and their context diversification, such as presence of different distractors and their arrangement, is often insufficient and done arbitrarily based on unspecified criteria. The evaluations focus on the success of completing tasks \cite{james2020rlbench,mees2022calvin}, efficiency \cite{mandi2024roco,srivastava2022behavior}, or how well a policy reasons about the environment \cite{yang2024octopus,jiang2022vima}. A recent study \cite{Pumacay_RSS_24} looks at the impact of environment clutter, considering the role of distractors, targets' characteristics, lighting, camera pose, and background texture. The authors show that each of these factors influences manipulation policies. This study, however, has two shortcomings. Distractors are treated as a single factor, therefore impact of their properties, quantities or arrangement in the scene on the performance of policies cannot be quantified. Moreover, the environmental factors that contribute to clutter are considered in isolation, as a result their compounding effects are omitted. For instance, the effect of camera pose or lighting can vary with different types or sizes of objects.

To address these limitations, we propose a novel evaluation protocol for robotic manipulation policies in diverse contexts. We approach the problem from a psychophysical perspective and make use of a unified clutter measure \cite{rosenholtz2007measuring} to characterize experimental scenarios populated with various types of distractors. We design scenes with diverse set of distractors in different quantities and include partial occlusion while ensuring the reachability of the target object without the need for rearranging the scene. We conduct large-scale evaluation in a hyper-realistic simulator, SIMPLER \cite{li2024evaluating} and a real-world environment by deploying 5 state-of-the-art  vision-language-action (VLA) manipulation policies to perform various manipulation tasks. We perform in-depth analysis of the policies performance in order to highlight the impact of clutter, identify strengths and weaknesses of the policies, and measure the differences among them. We show how our clutter measure can be used as an indicator of expected performance, and take a closer look at some of the individual effects of distractors. At the end, we examine training data enhancement strategy by finetuning a VLA policy on real-world distraction data to determine how well it can lower the negative effect of visual clutter.


\section{Related Works}

\textbf{Distractors in visual scene understanding.} In the vision literature,  distractors are defined as stimuli that do not contribute to task goal completion (i.e. they do not serve any functional purpose but increase the complexity) \cite{liesefeld2024terms}. Distractors come in many forms depending on their saliency, appearance, and similarity to the target. The literature on the effects of distractors is extensive. Numerous psychology studies investigated the effects of different types of distractors in visual search \cite{olk2018measuring, petilli2020distractor} and the attention mechanisms that suppress their impact \cite{tsotsos1995modeling,chelazzi2019getting} have been widely studied. In computer vision, methods have been proposed to alleviate the problems caused by distractors, e.g. category confusion in object detection \cite{li2021few,liu2022open}, target confusion with distractors due to their similarity \cite{zhu2018distractor}, and occlusion \cite{zhong2021towards} in object tracking.

In robotics, distractors have also been shown to influence performance. For instance, studies on autonomous driving based on a recent benchmark \cite{sun2024causalagents} show that perturbing distractors (here referred to as irrelevant or non-causal agents) can significantly hinder predictive performance of models and to counter such an effect a causal learning paradigm should be adopted \cite{pourkeshavarz2024cadet, ahmadi2024curb}. salient distractors are also show to negatively impact object search \cite{rasouli2016sensor,rasouli2020attention,rasouli2014visual} and lead to inaccuracies in localization and navigation \cite{mendez2021improving}.  In robotic manipulation, the presence of distractors in clutter has been shown to impact the recognition capability of the robot and pose challenges to grasping and manipulating the target \cite{dipalo2024kat, kim2020using, kim2021transformer, Samani_2024_persistent, kasaei2024simultaneous, tang2023selective, Ummadisingu2022food}. In some cases, distractors can also have a compounding effect on action generation in a given context. For example, in a small study, the authors of \cite{karamcheti2023voltron} show that changing the distractors by swapping them with similar objects of different color or different objects can drastically lower policy success rate, by up to 50\%, across various manipulation tasks.


\textbf{Robot simulators.} Owing to advancements in robotic simulation engines, large-scale evaluation of policies in simulated worlds has become wide-spread \cite{zhu2020robosuite, gupta2020corl, zeng2021transporter, jiang2022vima}. These simulators allow evaluation of more complex tasks, such as food preparation \cite{mandi2024roco} and furniture assembly \cite{lee2021ikea}, collaborative tasks, such as inspection and handover \cite{thumm2024human}, and tasks that require effective spatial reasoning \cite{liu2023libero}.  To minimize the sim-to-real gap, SIMPLER \cite{li2024evaluating} offers hybrid simulated scenes, in which background scenes are created from real data, objects, and robot arms, are post-processed by adding realistic textures. By comparing the performance of SOTA manipulation policies in real-world scenes and their simulated counterparts, the authors show that there is a high correlation between the performances of the robot in both environments. We therefore build our simulated evaluations within the SIMPLER environment. 

\textbf{Evaluations protocols for robotic manipulation.} The majority of manipulation benchmarks primarily focus on measuring the success rate on scenes that are often designed and populated in an unspecified manner. Works, such as \cite{james2020rlbench,mees2022calvin}, distractors are introduced into the environment, but their arrangement, types, or impact are not specified. Some works focus on the effect of distractors arrangement, whether they are similar \cite{mnyusiwalla2020bin} or distinct \cite{morgan2019benchmarking,bottarel2020graspa}. The shelving challenge \cite{eppner2016lessons}, added distractors (non-target objects) to the pick bins and if they are picked instead of the target, a penalty is applied. The authors of \cite{zheng2024robocas}, focus on the arrangement of background objects, placing them in a scattered, ordered (aligned with edges), or stacked configuration, without specifying the motivation behind such strategy. A more recent work \cite{Pumacay_RSS_24} analyzes the impact of environment factors, distractors, targets' characteristics, lighting, camera pose, and background texture on manipulation policies and concludes that each of these factors has different degrees of adverse effect. This study, however, does not characterize distractors in terms of their properties, quantities, or arrangement in the scene, hence their impact on the performance of policies cannot be quantified. Additionally, factors that contribute to clutter are considered in isolation, as a result, their compounding effects are unknown. These factors are not independent because, for instance, effect of lighting condition can be different on objects with different appearances, or camera pose impact can vary depending on the size and the arrangement of objects. In this work we propose a novel evaluation approach from a psychophysical perspective. We employ a unified measure of clutter that captures both characteristics and arrangement of distractors as well as environmental conditions. We systematically generate scenarios in order to quantify the impact of clutter and identify strengths and vulnerabilities of manipulation policies. 
\begin{figure*}[t]
    \centering
    \includegraphics[width=1\textwidth]{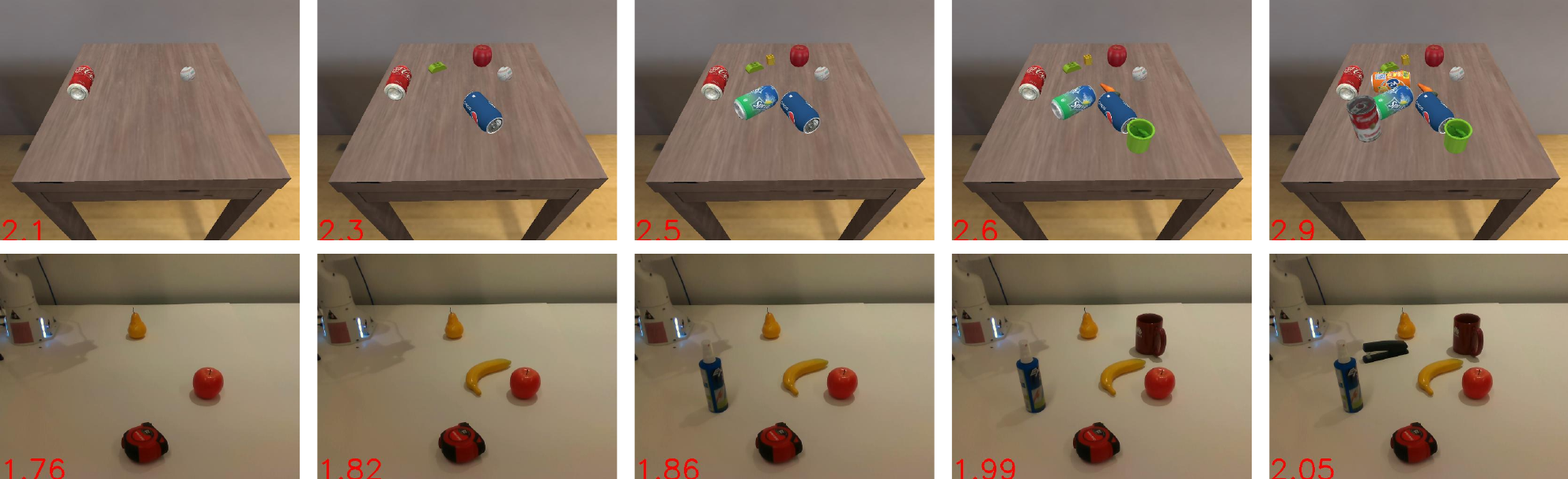}
\caption{Examples of various synthetic (top) and real (bottom) cluttered scenes with corresponding DvFC values.}
\label{fig:DvFC_cal}
\end{figure*}

\section{Evaluation Protocol}
\subsection{Problem formulation}
\textbf{Robot skills.} We address robotic manipulation problem in which the objective is to learn to generate actions given a visual observation and language instruction. Robot actions are in the form of 6-DOF gripper pose in the 3D space. For tasks, we consider core robotics skills and adopt the defaults tasks from \cite{li2024evaluating}, namely pick (lift) object, move object next to another object, stack objects, and place (smaller) objects on another (bigger) one. Here, the robot should detect the target accurately among the distractors, reach and grasp the target, potentially in a confined environment, and pick, move, and place it on the target location while avoiding collisions. Other tasks, such as pushing, pulling, tossing, etc. can also be negatively impacted by distractors.  We speculate that if clutter impacts the selected core abilities, by extension it can affect other tasks as well.

\textbf{Tasks.} As discussed earlier, our goal is to examine the impact of visual clutter on robotic manipulation tasks. To achieve this, there are several factors to consider \cite{semizer2023visual}, including the number of objects (set size), distractor similarity to the target object, their arrangement (whether they are well-aligned or placed randomly), and environmental factors, such as background texture, lighting, etc. Each factor has several sub-factors to consider. For instance, object similarity can be in terms of geometry or color. A dense placement can limit the operational space of the robot by reducing the affordance or can cause visual occlusion and therefore impact the detectability of the target. In addition, there is a interaction effect that should be factored in. For instance, a tall object placed next to a short target can significantly limit the reachability of the target, whereas this is not the case if they are spaced sufficiently apart. 

Systematically permuting all these factors to generate a complete set of scenarios is infeasible, if not impossible. Thus, we  approach this problem from a psychophysical perspective and take a holistic view of clutter. We use the degree of clutter as a measure of scene complexity due to distractors and fix other environmental factors. 

\subsection{Clutter measure} Our goal is to define a measure that characterizes our evaluation scenarios. This would allow us to quantify the impact of different degrees of scene clutter on the performance of the policies. Unlike the past works that treatedfactors of scene clutter individually \cite{Pumacay_RSS_24}, we use a unified measure to account for compounding effect of different scene elements and minimize factor bias in our evaluation. 

There are different methods of measuring visual clutter, often tied to certain applications, e.g. mapping \cite{frank1994multiple}, object visibility \cite{woodruff1998constant}, or arrangement \cite{mack2004computational}. We adopt a psychophysical metric, namely, the feature congestion measure (FCM)\cite{rosenholtz2007measuring}, that combines the covariances of scene color, contrast, and orientation at different scales, which effectively capture distractors quantity, spatial distribution and similarity to the target as well as environmental characteristics. This metric, however, is designed for 2D images and does not parse the operational complexity or fail to account for objects that are occluded. To remedy this, we propose a dual-view approach, in which we combine the measures from the robot view and the top-down view. In this way, we can measure both visual clutter from the policy point of view and operational complexity of the actions. We refer to our measure as dual-view feature congestion (DvFC). As shown in qualitative examples in Figure \ref{fig:DvFC_cal},  introducing diverse objects to the scenes increases the clutter level, and corresponding DvFC accordingly. Syntehtic scenes generally have higher DvFC values as the objects are placed more densely and the background is more textured compared to the real scenes.


\subsection{Scenario Generation}
 We sample from the base scenarios in SIMPLER, randomly select 1--12 objects from our distractor set of 61 YCB objects\cite{calli2015benchmarking}, and randomly place them within the operational space of the robot. For spacing, we consider a minimum $\delta$ gap between the objects to avoid stacking or piling of the objects. The space around the target is also constrained to minimize the impact on grasping affordance. Lastly, for each generated scene, we compute DvFC.

\textbf{Scenario sampling.} To ensure feasibility of actions in the scenes, we first discard all scenarios in which target objects are significantly visually occluded (more than 50\%) or have no grasping affordance. We then group the remaining scenes according to their DvFC scores into N bins and uniformly sample from them.

\section{Experiments}
\label{sec:result}
We pursue two key objectives: measuring the impact of distractors on the performance of policies and observing the differences between policies in handling challenging scenarios. More specifically, we seek answers to the following questions: 1) How do distractors impact the success of manipulation policies? 2) Do policies perform similarly in cluttered scenes? 3) Is scene clutter measure a good estimator of policy's performance? 4) What aspects of the resulting scene clutter have the greatest impact on performance?

\begin{table*}[t]
\centering
\caption{The average performance of manipulation policies in cluttered scenes. SR$_{base}$ refers to the performance on original scenarios, SR$_{noocc}$ and SR$_{occ}$ refer to success rate in scenarios without and with visual occlusion of the target object. Arrows, $\uparrow$ and $\downarrow$, indicate whether higher or lower values are better.}
\resizebox{0.8\textwidth}{!}{
 \begin{threeparttable}[b]
\begin{tabular}{l|c|ccccccc}
\textbf{Policy}     & \textbf{SR$_{base}$$\uparrow$} & \textbf{SR$\uparrow$} & \textbf{SR$_{noocc}$$\uparrow$} & \textbf{SR$_{occ}$$\uparrow$} & \textbf{h-SR$\uparrow$} & \textbf{CR$\downarrow$} & \textbf{GFR$\downarrow$} &\textbf{ER$\downarrow$} \\ \hline
\textbf{Octo*}       &     0.173            & 0.117       & 0.104                & 0.013            & 0.011         & 0.867       &0.439& 0.490      \\\hline
\textbf{OpenVLA**}    &         0.533          & 0.262       & 0.170                & 0.920            & 0.070         & 0.590       & 0.282& 0.517      \\\hline
\textbf{SpatialVLA} &              0.587     & 0.240       & 0.169                & 0.072            & 0.084         &\textbf{0.684}       & 0.444&0.368      \\
\textbf{$\pi_0$}      &            0.711       & 0.470       & 0.346                & 0.124            & 0.093         & 0.840       & \textbf{0.287}&0.470      \\
\textbf{CogACT}     &        \textbf{0.743}           & \textbf{0.480}      & \textbf{0.348}                & \textbf{0.132}           & \textbf{0.102}        & 0.872       & 0.395& \textbf{0.367}     
\end{tabular}
\begin{tablenotes}
   \item \tiny{*Evaluated only on Bridge scenarios. **Evaluated only on Fractal scenarios.}
 \end{tablenotes}
 \end{threeparttable}
 }
\vspace{-0.3cm}
\label{tab:average_res}
\end{table*}
\subsection{Experimental Setup}
\textbf{Scenarios.} We use the following six skills from SIMPLER default categories, namely \textbf{Move} near, \textbf{Stack} cube, \textbf{Pick coke}, and 3 from the Bridge pick-and-place tasks, \textbf{Put spoon}, \textbf{Put eggplant}, and \textbf{Put carrot} (see \cite{li2024evaluating} for more details). In each scenario, we systematically add distractors randomely sampled from 61 YCB objects \cite{calli2015benchmarking} to the environment. Distractors that would make the instruction ambiguous are removed from distractor selection. The placement of each distractor on the tabletop is random with two constraints to ensure that the target can be grasped. First, we consider a distance threshold between the distractors and the target to ensure separation between the target and the nearby objects. Second, a condition is set to limit the visual occlusion of the target (from the default perspective of the robot) to a maximum of 50\%. In total, we generated 6000 scenarios.
 
\textbf{Models.} We follow SIMPLER and evaluate five state-of-the-art vision-language-action (VLA) models, including Octo \cite{mees2024octo} (trained on Bridge dataset\cite{walke2023bridgedata}), OpenVLA \cite{kim2024openvla} (trained on Fractal dataset\cite{brohan2022rt}), and CogACT \cite{li2024CogACT}, $\pi_0$ \cite{black2410pi0}, and SpatialVLA \cite{qu2025spatialvla}, which are trained on both Bridge and Fractal\cite{walke2023bridgedata,brohan2022rt}. Note that since Octo and OpenVLA are only trained on one of the two datasets, we report their results only for the corresponding sets as a reference while using the three other policies for our main evaluation.

\textbf{Metrics}
We follow the common protocols and report success rate (SR) as the primary metric \cite{li2024evaluating,gupta2020corl,zeng2021transporter}. SR is computed as the percentage of tasks completed even if collision occurred. We also consider hard SR (h-SR), which measures success cases without any collisions. In addition, we report on collision rate (CR), measured as the percentage of the scenarios in which a collision occurred. We consider any contact with a distractor as a collision. For failure analysis, we also report on grasp failure rate (GFR), that is the percentage of scenarios in which the robot fails to grasp the target. Lastly, we consider efficiency rate (ER), in terms of the number of steps it takes the robot to complete its task, normalized by the total number of steps allowed in the given scenario. Our initial observation is that clutter can potentially create visual confusion leading the robot to reach for incorrect objects before grasping the target, hence the number of steps for completion can potentially be increased.


\subsection{Evaluation in Simulated World}

\textbf{Distractors significantly hinder the performance of manipulation policies. } We begin evaluating the policies by averaging their performance across all newly generated scenarios. Here, in addition to the overall result, we report the success rate for scenes with or without visual occlusion of the target. Note that Octo and OpenVLA are trained on a subset of data and evaluated on the corresponding tasks from those subsets, hence are only included as a reference. 

As shown in Table \ref{tab:average_res}, increasing scene clutter by the addition of distractors significantly reduces the performance of all policies. $\pi_0$ and CogACT generally have a higher degree of success, achieving approximately 50\% in SR. The success rate across visual occlusion scenarios is significantly lower, pointing to the potential for target confusion and collision. It is also interesting to note that the efficiency of the policies varies significantly. While SpatialVLA and CogACT achieved the best efficiency, $\pi_0$ is 10\% less efficient comparatively even though its SR is only 1\% below CogACT.  

High CR values indicate that the policies lack effective obstacle avoidance mechanisms. An exception is SpatialVLA, which collided 16\% less compared to the next best model $\pi_0$. This is also reflected in h-SR metric where the gap between SpatialVLA and the other two  policies is much lower, despite the much larger differences in SR. This might be attributed to the fact that SpatialVLA is optimized for a better spatial understanding of the environment. In terms of GFR, $\pi_0$ clearly stands out (compared to models evaluated on all scenarios), indicating  the robustness of this policy's grasping affordance estimation.

 \begin{figure}
      \centering
      \includegraphics[width=0.6\columnwidth]{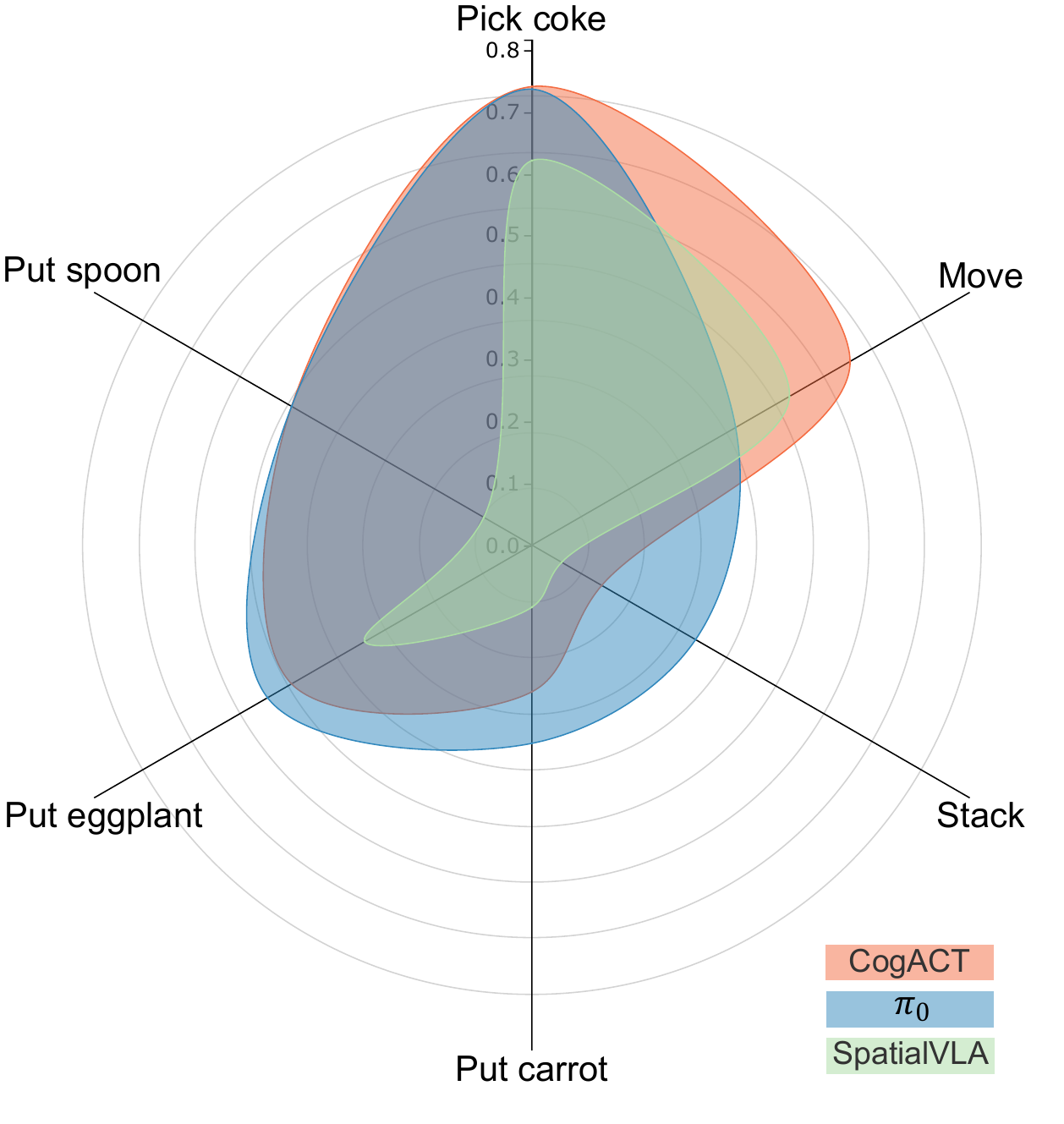}
    \caption{Per-task success rate of the policies. Each axis of the radar diagram shows one of the 6 core tasks.}
    \label{fig:radar_per_task}
\end{figure}

 \begin{figure}
      \centering
      \includegraphics[width=0.6\columnwidth]{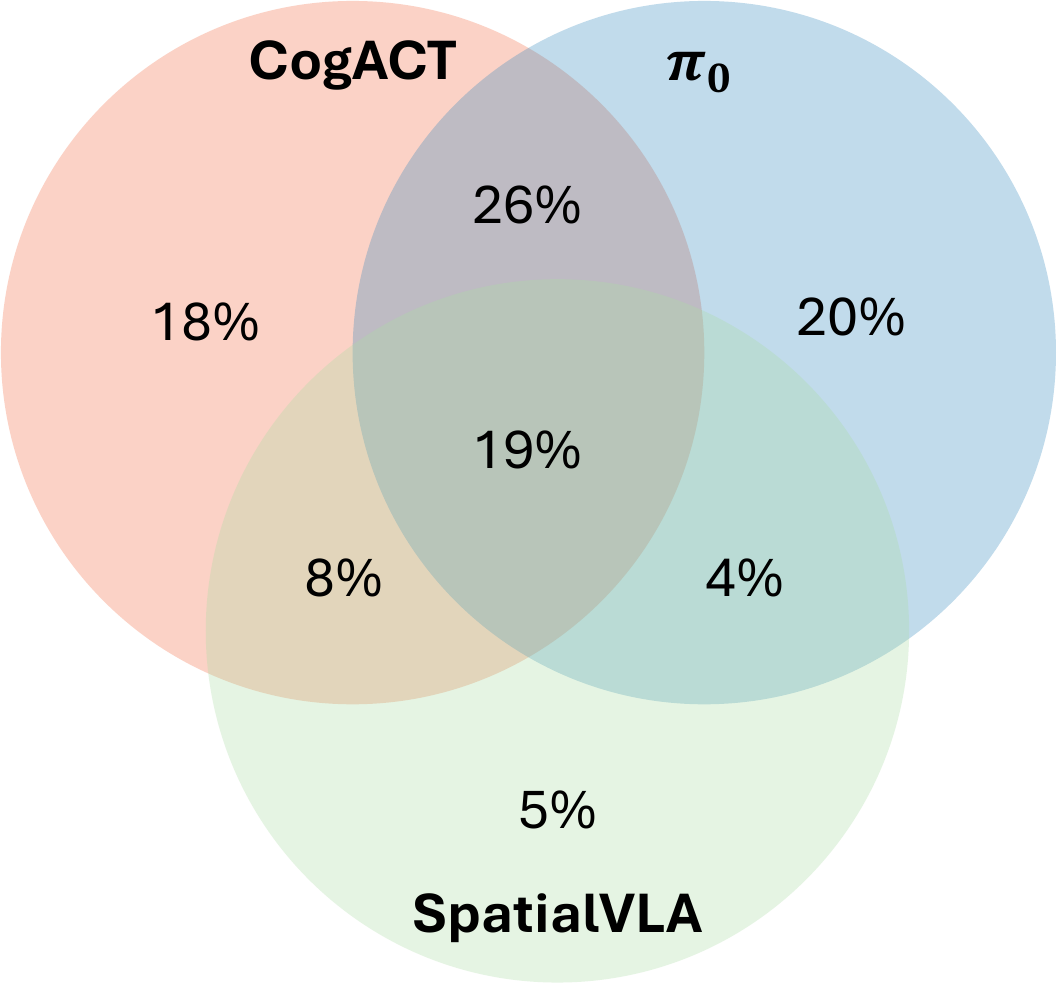}
    \caption{The Venn diagram of the success scenarios of the  policies. The numbers show the percentage of total success scenarios combined for the policies.}
    \label{fig:venn_average} \vspace{-0.4cm}
\end{figure}

 \begin{figure*}
      \centering
      \includegraphics[width=1\textwidth]{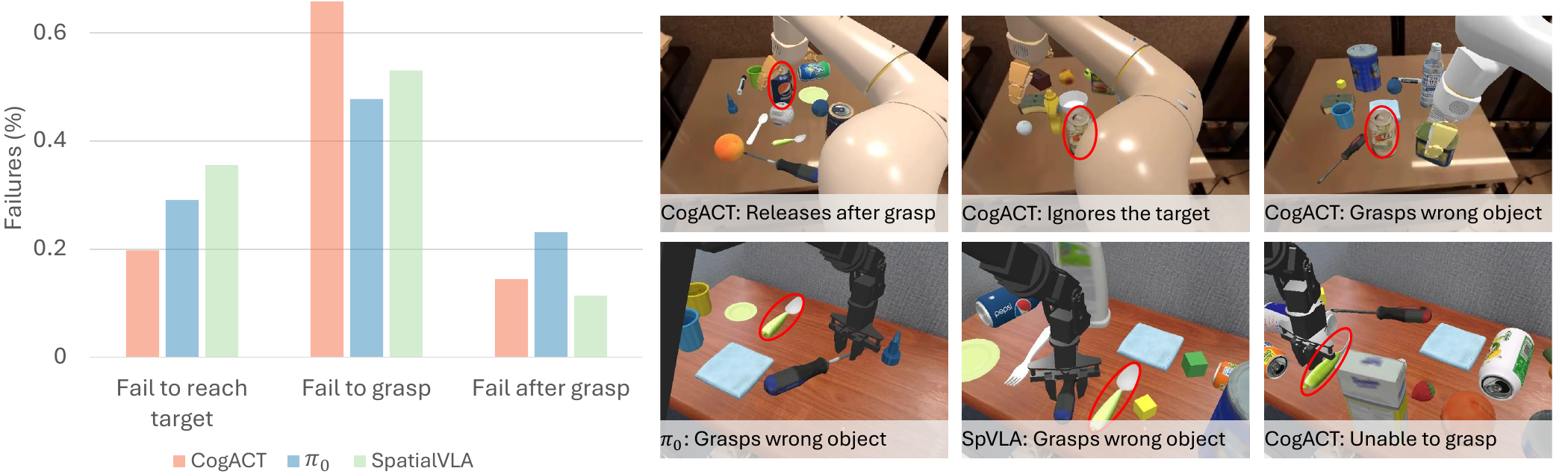}
    \caption{(\textbf{left}) Percentage of failures for the policies. Values for each policy are normalized to sum to 1. Lower values are better. (\textbf{right}) Qualitative examples of failure cases. Targets are identified with red ovals.}
    \label{fig:sources_failure}
\end{figure*}

\begin{figure}
    \centering
    \includegraphics[width=1\linewidth]{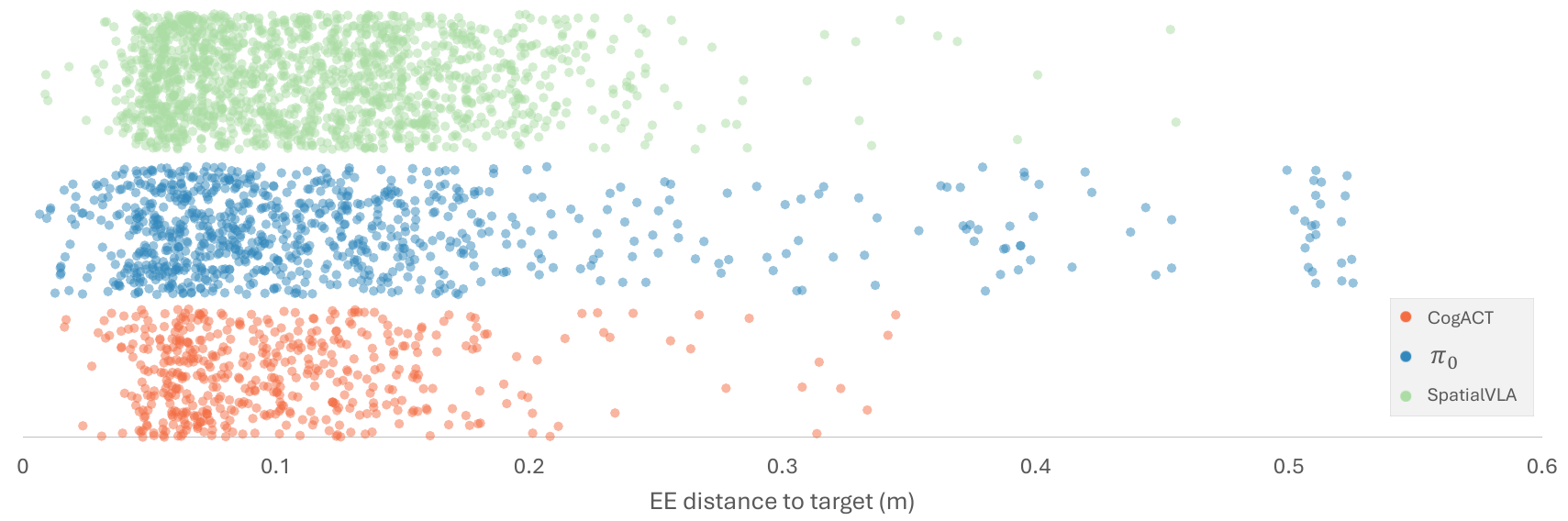}
    \caption{Distribution of the per-scenario \textbf{Fail to reach target} failure cases. Each circle indicates the closest the end effector (EE) got to the target prior to failure. Circles with the same Y coordinate are randomly perturbed for displaying purposes.}
    \label{fig:dist_to_target}
\end{figure}

\textbf{Policies are affected differently by distractors.}  As shown in per-task performance in Figure \ref{fig:radar_per_task}, the peak performance for all models is on the simplest task, Pick coke (as involves only lifting the target) but the performance on other tasks varies. SpatialVLA and CogACT perform better on Move, whereas $\pi_0$ performs better on Stack and two out of the three Put tasks. Overall, $\pi_0$ demonstrates a more balanced performance compared to the other two policies. 

Despite having a significant overlap on per-task performance, policies are not necessarily successful in the same scenarios. According to Figure \ref{fig:venn_average}, CogACT shares only about 45\% of its successes with $\pi_0$, even though both have a similar average SR. Each policy also succeeds in a large subset of scenarios in which the other two policies fail. This indicates that policies have complementary performance despite having similar architectures and being trained on similar data. In fact, combined, the policies achieve approximately 67\% SR in all new scenarios. 
\begin{figure*}[t]
    \centering
         \begin{subfigure}{1\linewidth}
         \includegraphics[width=1\linewidth]{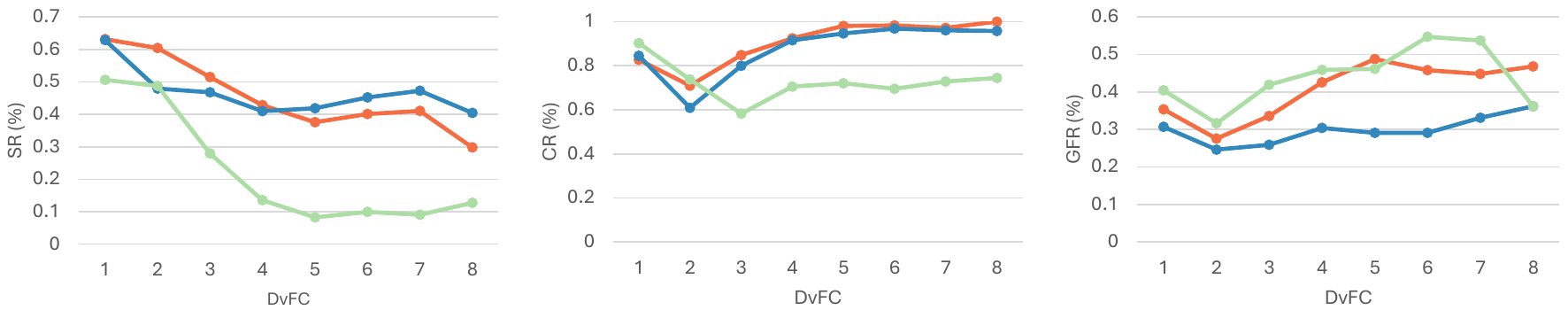}
     \caption{Visual clutter}
     \label{fig:DvFC_vs_sr}
     \end{subfigure}
     \begin{subfigure}{0.33\linewidth}
    \includegraphics[width=1\textwidth]{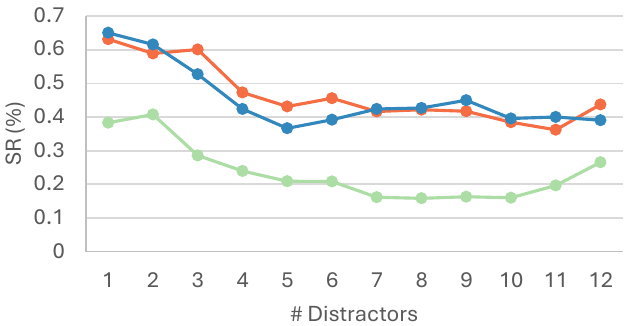}
    \caption{Set size}
    \label{fig:set_size_sr}
\end{subfigure}
\hspace{0.5cm}
\begin{subfigure}{0.46\linewidth}
    \includegraphics[width=1\textwidth]{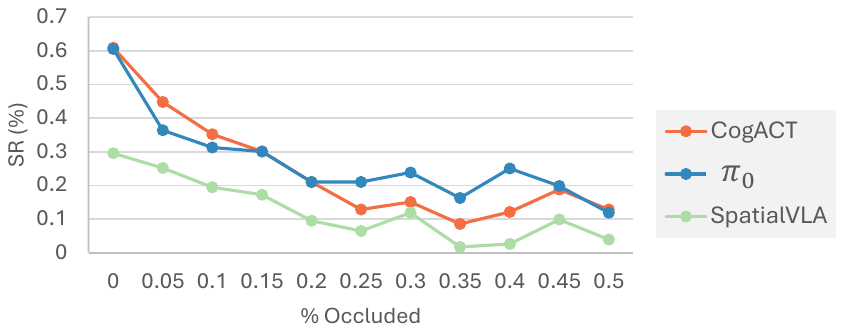}
    \caption{Visual occlusion}
    \label{fig:occlusion_sr}
\end{subfigure}
\caption{a) Performance of policies at different levels of visual clutter. For SR higher is better and for CR and GFR lower.  Policies success rate b) with different number of distractors and c) at different collision rates.}\vspace{-0.5cm}
\label{fig:clutter_vs_sr}
\end{figure*}

Besides differences in success scenarios, the policies' failures also vary. We show that by considering the following 3 stages of operation: \textbf{Fail to reach target}---indicating whether the gripper reaches the target object; \textbf{Fail to grasp}---inability to grab the target; and \textbf{Fail after grasp}---failures that occurred after grasping action. Note that the failure types are not double-counted, for instance, failure to reach target does not count towards failure to grasp or after grasp. 

The summary of failures is illustrated in Figure \ref{fig:sources_failure}. Here, the percentage of failures for each policy for each type of error is shown. Once again, there are noticeable differences among the policies. For example, CogACT is generally more successful in reaching the correct target object but lags behind on grasping, and $\pi_0$ is generally better at grasping the target but is worse at completing the task. 

According to the qualitative examples in Figure \ref{fig:sources_failure} the causes of these types of failures vary. For example, in Fail to reach target cases, besides collision, grasping wrong objects (distractors) due to visual confusion is prevalent. For instance, in the bottom left and middle scene the robot grasps the screwdriver instead of the spoon, or in the top right sample grasps the can of spam (which has a cuboid shape with blue color) instead of 7UP (which is cylindrical and green). In some other scenarios (top middle), the robot simply passes over the target (a can of 7UP) without reaching for it or repeatedly grasps and releases the can of Pepsi (top left) without lifting it for no apparent reason.

To further study the differences between the policies, we break down the \textbf{Fail to reach target} and summarize the results in Figure \ref{fig:dist_to_target}. Here, we consider the closest distance of the end-effector to the target prior to the failure.  As shown in the figure, CogACT is generally more successful in reaching the target as the points are more concentrated closer to the target location. However, the failure points of $\pi_0$ are more dispersed and extend beyond 0.5m from the target. This indicates that this policy is more vulnerable to target confusion and misidentification of objects.

\textbf{Clutter measure is a strong indicator of the expected policy performance.} We divide scenario DvFC values into 8 bins and compute the success rates of the policies for each bin. As shown in Figure \ref{fig:DvFC_vs_sr}, there is a general downward trend in performance as the DvFC value increases, although the rates of change are different for each policy. CogACT and $\pi_0$'s performance declines at the beginning and stabilizes with minor fluctuations in mid-range DvFC values before further decreasing at the end. SpatialVLA's performance, on the other hand, drops rapidly early on and reaches the minimum SR midway. This means SpatialVLA is more significantly impacted by clutter compared to the other polices. Overall, $\pi_0$ shows more stability at higher clutter. 

The differences  are also apparent in error rates. As shown in Figure \ref{fig:DvFC_vs_sr}, while $\pi_0$ and CogACT have high collision rates, SpatialVLA shows a more stable performance which is aligned with its better average CR. The variation in grasp failure rate (GFR), however, is much higher.  $\pi_0$ generally performs better as its GFR rises slowly,  increasing by 10\% across different clutter values whereas other policies degrade by more than 20\%. The final drop in GFR of SpatialVLA can be due to general lower success rate of this policy in highly cluttered scenes.

\textbf{Different aspects of distractors and clutter have different negative impact.} Although all policies showed a downward performance trend (see Figure \ref{fig:DvFC_vs_sr}), there were some irregularities at the higher DvFC values. This is due to the fact that there are many factors at play in a cluttered scene, including congestion, distractors' properties (such as shape, color, size), set size, visual occlusion, etc. To examine their impact, we further group the scenarios according to two of these factors, namely set size (i.e. the number of distractors) and visual occlusion. According to Figure \ref{fig:set_size_sr}, we can see that adding distractors to the scene negatively affects the overall performance, although after 5 distractors the performance of all policies remains stable. In the case of target visual occlusion, as shown in Figure \ref{fig:occlusion_sr}, we observe a very different trend. The initial performance decline for all policies is drastic before stabilizing at roughly 20\% occlusion. However, the rate of decline varies. As an example,  while $\pi_0$ is affected more at the beginning compared to CogACT, it stabilizes at a higher SR value, indicating that this model is more robust to visual occlusion. Among the main policies, SpatialVLA is the most vulnerable, dropping to 0\% SR at the highest occlusion level.

\begin{figure*}
    \centering
    \includegraphics[width=1\linewidth]{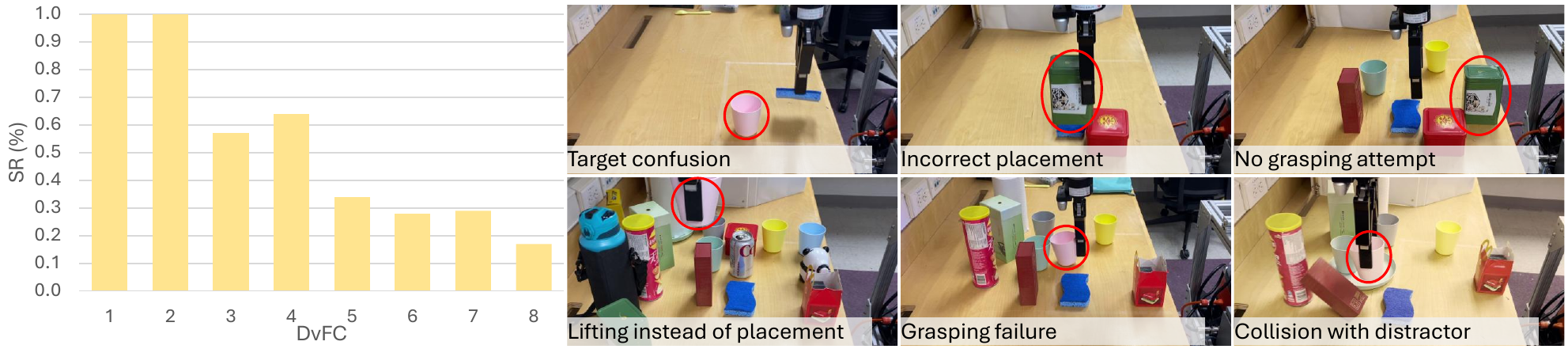}
    \caption{(\textbf{left}) Experimental results on real-world cluttered scenes. 1-8 indicates the bins according to DvFC measure. (\textbf{right}) Qualitative examples of failures in real world. Targets are identified with red ovals.}
    \label{fig:real_exp}\vspace{-0.5cm}
\end{figure*}
\subsection{Evaluation in Real World}
We validate our findings by replicating our experimental setup in real world. However, due to prohibitive amount of time needed to generate a large number of scenarios, we resort to more sparse sampling of distractors. More specifically, we consider \textbf{6 variations}, with 0, 1, 2, 4, 8, and 16 distractors. Due to the larger operational space of our robot, we chose a larger number of distractors. We consider the four core skills, namely pick, move, stack, and put, as in our simulated experiments and create 9 variations of each setup totaling 216 scenarios. 

We choose $\pi_0$  as it demonstrated more balanced performance across all skills. We finetuned the policy using 42 samples we collected for each skills. In this data we only placed the target object(s) without any distractors. For the sake of brevity, we only report on the overall success rate of the policy over all tasks. All experiments are done using a UR5e robot manipulator. Similar to the simulated experiments, we break down the scenarios into 8 bins according to their DvFC measure.

As shown in Figure \ref{fig:real_exp}, once again we observe a downward trend in performance as the clutter level increases. As in the simulation experiments (see Figure \ref{fig:clutter_vs_sr}), we can see some small fluctuations, e.g. from bin 3 to 4, which can be due to the model uncertainty or individual factors causing the clutter. Further, the impact of clutter intensifies as the complexity of the task increases.  The qualitative samples of failures are shown in the figure. Similar to the synthetic experiments, target confusion is prevalent, occurring even by adding as few as one distractor to the scene. 

\begin{figure}
    \centering
    \includegraphics[width=0.8\linewidth]{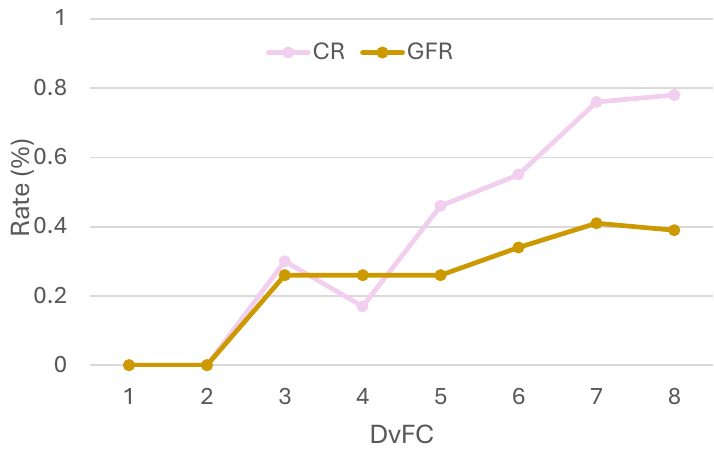}\vspace{-0.3cm}
    \caption{Collision rate (CR) and grasp failure rate (GFR) versus visual clutter measure in real-world experiments. Lower values are better.}
    \label{fig:err_real}\vspace{-0.5cm}
\end{figure}

$\pi_0$'s failures in real-world (see Figure \ref{fig:err_real}) also exhibit similar trends to synthetic results. Reaching clutter level of 3, we observe a significant increase in CR and GFR. CR generally increases more rapidly whereas GFR despite its initial increase stabilizes, confirming our findings from the simulated experiments.

\section{Does Data Help with Clutter Scenes?}
Thus far we showed that clutter caused by distractors significantly affect the performance of the policies. There are a number of ways one can approach this problem, ranging from explicit scene reasoning via architectural changes to data enhancement. Since the focus of our study is on VLAs, we employ the latter approach and examine the effectiveness of data in real-world scenarios.

We follow our evaluation protocol and generate data by creating scenes with different numbers of distracting objects. The distracting objects and arrangements are randomized to differ from our test scenarios used earlier. For each of the four skills, besides the base samples, we collected 45 scenarios with distractors present (9 arrangements for each of 5 levels of distractors as discussed before). Using the new data, we finetuned the base $\pi_0$ model and evaluated the policy on our test scenarios. 
\begin{figure}
    \centering
    \includegraphics[width=0.8\linewidth]{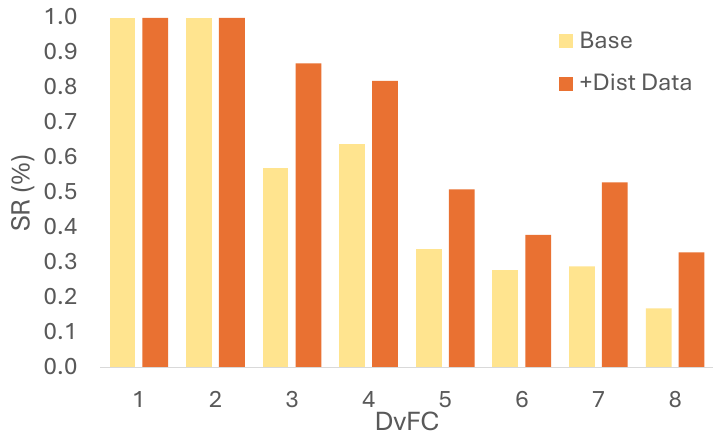}
    \caption{Comparison of the policy finetuned on base data and distractor data.}
    \label{fig:real_data_exp} \vspace{-0.3cm}
\end{figure}

As shown in Figure \ref{fig:real_data_exp}, enhancing tuning data with distractor samples increases the robustness of the policy to clutter. The improvement, however, is not consistent across all metrics. According to Table \ref{table:fintune_dist}), although success rate increases by 18\%, other metrics show less of an improvement, e.g. only 6\% in GFR. It is foreseeable that additional data would help further improve the performance. However, scaling the data as the only solution is questionable because the gain is not very significant even in our controlled environment with fixed types of distractors.

\begin{table}[]
\caption{Performance comparison of the policy finetuned with base and distractor data. Arrows indicate whether higher or lower values are better.}\label{table:fintune_dist}
\centering
\resizebox{0.8\linewidth}{!}{
\begin{tabular}{l|cccc}
                     & \textbf{SR$\uparrow$} & \textbf{h-SR$\uparrow$} & \textbf{CR$\downarrow$} & \textbf{GFR$\downarrow$} \\ \hline
\textbf{Base}        & 0.47        & 0.33          & 0.38        & 0.29         \\
\textbf{+Dist. data} & \textbf{0.65}      & \textbf{0.48}          & \textbf{0.27}    & \textbf{0.23}        
\end{tabular}}\vspace{-0.6cm}
\end{table}


\section{Conclusion}
\label{sec:conclusion}
In this work, we conducted a study on the impact of visual clutter on robotic manipulation policies from a psychophysical perspective.  Our findings, from both simulated and real-world experiments, demonstrated significant negative impact of clutter on performance. Our analysis indicates that averaging success of the policies across all scenarios is not sufficient to effectively highlight the strengths and weaknesses of the policies.  Despite having similar average performance, the policies have complementary strengths, were affected by environmental factors differently, and were vulnerable to different types of failures. We showed that visual clutter measure was an effective predictor for the performance of policies. Lastly, we examined the role of data enhancement as a way of improving robustness to clutter. Although improvements were observed, the gain across all metrics was not substantial, even in our controlled environments with limited variability. Our study points to the need for better examination and evaluation of policies across different scenarios and studying alternative approaches for clutter handling, besides data scaling.

\bibliographystyle{IEEEtran}
\bibliography{refs}  

\end{document}